\documentclass[journal]{IEEEtran}
\usepackage{algorithmic}
\usepackage{algorithm}
\usepackage{array}
\usepackage[caption=false,font=normalsize,labelfont=sf,textfont=sf]{subfig}
\usepackage{textcomp}
\usepackage{stfloats}
\usepackage{url}
\usepackage{cite}
\newcommand{\myparagraph}[1]{\vspace{0.1em}\noindent\textbf{#1}}
\usepackage[colorlinks,hyperindex,breaklinks]{hyperref}
\newcommand{\ie}{\textit{i}.\textit{e}.}
\newcommand{\eg}{\textit{e}.\textit{g}.}

\usepackage{pifont}
\newcommand{\cmark}{\ding{51}}%
\newcommand{\xmark}{\ding{55}}%
\usepackage{graphicx}
\usepackage{colortbl}
\usepackage{arydshln}
\usepackage{multirow}
\usepackage{multicol}
\usepackage{mathrsfs}
\usepackage{bm}
\usepackage{amsfonts}
\usepackage{amssymb} 
\usepackage{ragged2e}
\newcommand\etc{etc\@ifnextchar.{}{.\@}}
\usepackage{amsmath}
\usepackage{amsthm}
\usepackage{verbatim}
\usepackage{latexsym}
\usepackage{silence}

\usepackage{booktabs}

\usepackage{makecell}
\def\BibTeX{{\rm B\kern-.05em{\sc i\kern-.025em b}\kern-.08em
    T\kern-.1667em\lower.7ex\hbox{E}\kern-.125emX}}
\usepackage{balance}
\begin{document}
\title{Coupling Global Context and Local Contents for Weakly-Supervised Semantic Segmentation}
\author{Chunyan~Wang,
        Dong~Zhang,~\IEEEmembership{Member,~IEEE},
        Liyan~Zhang, 
        Jinhui~Tang,~\IEEEmembership{Senior Member,~IEEE}
\thanks{
This work was supported in part by the National Key Research and Development Program of China under Grant 2018AAA0102002, the National Natural Science Foundation of China under Grants 61925204 and 62172212.
}
\thanks{C. Wang, D. Zhang and J. Tang are with the School of Computer Science and Engineering, Nanjing University of Science and Technology, Nanjing 210094, China. E-mail: \{carrie\_yan, dongzhang, jinhuitang\}@njust.edu.cn.}
\thanks{L. Zhang is with the College of Computer Science and Technology, Nanjing University of Aeronautics and Astronautics, MIIT Key Laboratory of Pattern Analysis and Machine Intelligence, Collaborative Innovation Center of Novel Software Technology and Industrialization, Nanjing 211106, China. E-mail: zhangliyan@nuaa.edu.cn.}
\thanks{Corresponding author: Liyan~Zhang.}
}
\markboth{IEEE Transactions on Neural Networks and Learning Systems}%
{Shell \MakeLowercase{\textit{et al.}}: Bare Demo of IEEEtran.cls for IEEE Journals}
\maketitle
\begin{abstract}
Thanks to the advantages of the friendly annotations and the satisfactory performance, Weakly-Supervised Semantic Segmentation (WSSS) approaches have been extensively studied. Recently, the single-stage WSSS was awakened to alleviate problems of the expensive computational costs and the complicated training procedures in multi-stage WSSS. However, results of such an immature model suffer from problems of \emph{background incompleteness} and \emph{object incompleteness}. We empirically find that they are caused by the insufficiency of the global object context and the lack of the local regional contents, respectively. Under these observations, we propose a single-stage WSSS model with only the image-level class label supervisions, termed as \textbf{W}eakly-\textbf{S}upervised \textbf{F}eature \textbf{C}oupling \textbf{N}etwork (\textbf{WS-FCN}), which can capture the multi-scale context formed from the adjacent feature grids, and encode the fine-grained spatial information from the low-level features into the high-level ones. Specifically, a flexible context aggregation module is proposed to capture the global object context in different granular spaces. Besides, a semantically consistent feature fusion module is proposed in a bottom-up parameter-learnable fashion to aggregate the fine-grained local contents. Based on these two modules, \textbf{WS-FCN} lies in a self-supervised end-to-end training fashion. Extensive experimental results on the challenging PASCAL VOC 2012 and MS COCO 2014 demonstrate the effectiveness and efficiency of \textbf{WS-FCN}, which can achieve state-of-the-art results by $65.02\%$ and $64.22\%$ mIoU on PASCAL VOC 2012 \emph{val} set and \emph{test} set, $34.12\%$ mIoU on MS COCO 2014 \emph{val} set, respectively. The code and weight have been released at:~\href{https://github.com/ChunyanWang1/ws-fcn}{WS-FCN}.
\end{abstract}
\begin{IEEEkeywords}
Weakly-supervised learning, single-stage semantic segmentation, context aggregation, feature fusion.
\end{IEEEkeywords}

\IEEEpeerreviewmaketitle
\section{Introduction}\label{intro}
\IEEEPARstart{S}{emantic} segmentation aims to predict each pixel of the input image with a unique category label, which is one of the most fundamental research topics in the computer vision domain~\cite{long2015fully}. In the recent past, this topic has been well studied and applied to a wide range of downstream applications, \eg, computer-aided medicine~\cite{zhang2022deep}, virtual reality~\cite{serrano2017movie}, and automatic drive~\cite{hu2020object}. However, for a fully-supervised semantic segmentation model, obtaining abundant and elaborate pixel-level annotations is unbearably expensive~\cite{zhang2020causal,ahn2019weakly,zhang2021self}.
Especially in the era of deep learning, we usually need a large number of training data so that the model can be optimized. Therefore, this problem will be particularly acute. To this end, Weakly-Supervised Semantic Segmentation (WSSS) has become the primary way to alleviate this problem. By ``weak'', refers to some easily obtained image labels, \eg, image-level class labels~\cite{ahn2018learning,araslanov2020single,zhang2020splitting}, bounding boxes~\cite{dai2015boxsup}, points~\cite{bearman2016s}, and scribbles~\cite{lin2016scribblesup}. Compared with the other three types, image-level class labels are the most challenging because they do not contain any object location information. In this work, we also use image-level class labels as the unique supervision.
\begin{figure}[tb]
\includegraphics[width=.48 \textwidth]{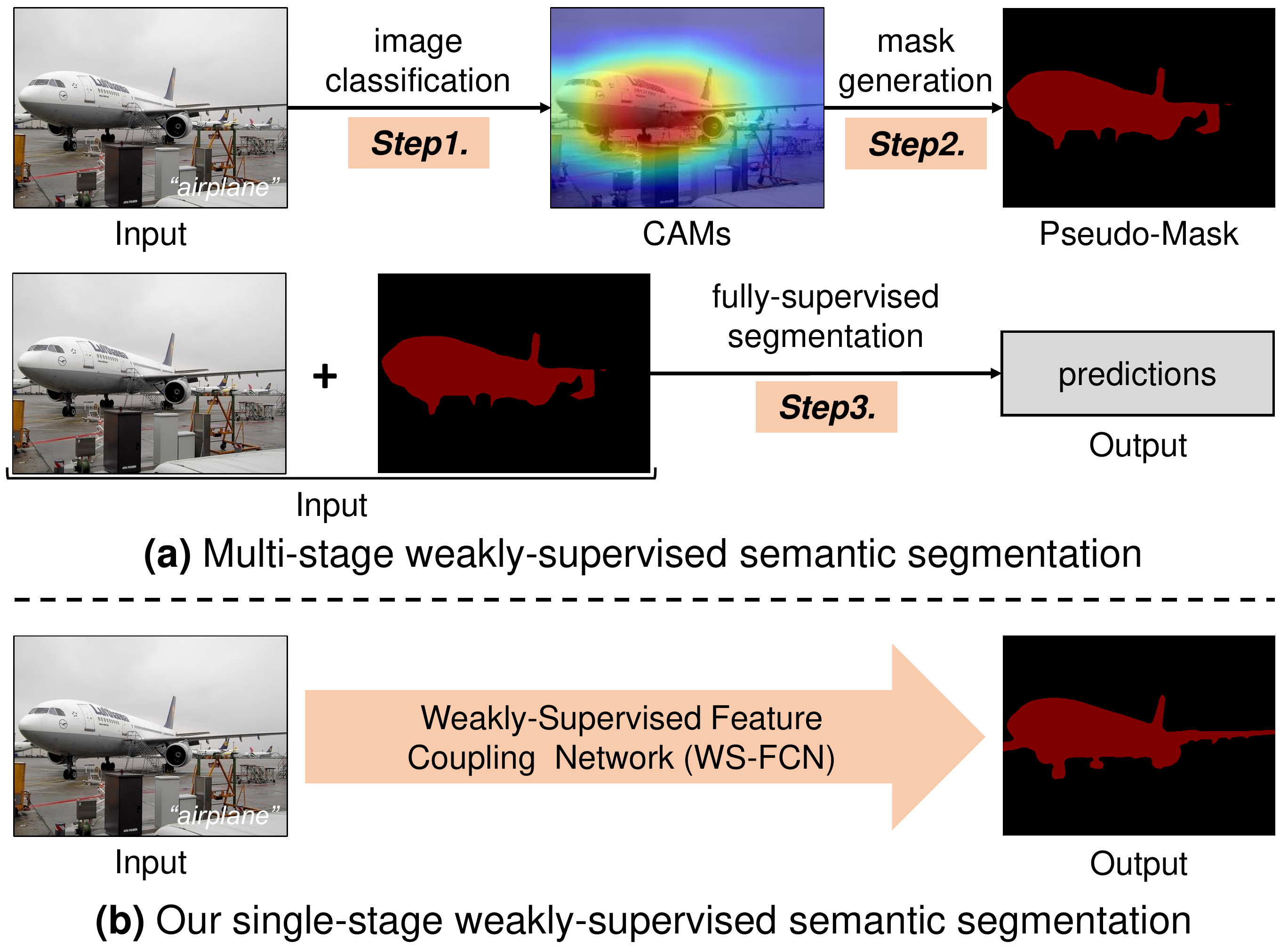}
\vspace{-2mm}
\caption{Illustration of the traditional multi-stage weakly-supervised semantic segmentation and our single-stage weakly-supervised semantic segmentation. Compared to the traditional one, our model has more concise training procedures and less computation costs while maintaining a competitive performance on accuracy. (Best viewed in colour.)}
\label{fig0}
\vspace{-2mm}
\end{figure}
\begin{figure*}[tb]
\includegraphics[width=.99\textwidth]{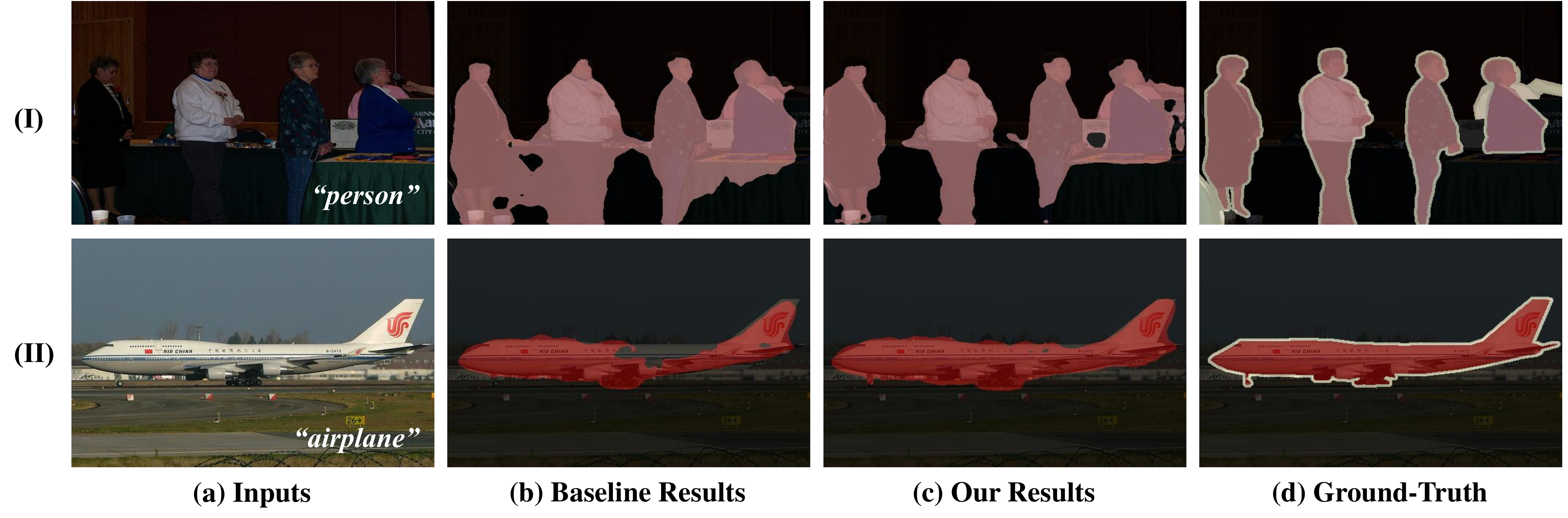}
\caption{Two basic problems in our baseline single-stage weakly-supervised semantic segmentation~\cite{araslanov2020single}, which are (\uppercase\expandafter{\romannumeral1}) \textbf{\emph{background incompleteness}} and (\uppercase\expandafter{\romannumeral2}) \textbf{\emph{object incompleteness}}. These two issues are usually mixed together to form more complicated situations in practice. Our proposed \textbf{WS-FCN} can accurately predict object boundaries while preserving abundant detail information. Samples are from the \emph{val} set of PASCAL VOC 2012~\cite{everingham2010pascal} dataset.}
\label{fig1}
\end{figure*}

The existing WSSS methods are mainly based on multiple processes, as illustrated in Figure~\ref{fig0}, which mainly consist of: {step-1}) localizing Classification Activation Maps (CAMs)~\cite{zhou2016learning}, {step-2}) generating pseudo-masks, and {step-3}) training a fully-supervised semantic segmentation model. 
Specifically, {step-1} first obtains the classification weights via a multi-label classification model and then uses the obtained weights to classify each pixel of training images to obtain CAMs. Based on this, {step-2} further removes some pixel regions with low classification confidence (below the given threshold) and then performs the ``seed expansion'' operation on the remaining regions to obtain pseudo-masks~\cite{kolesnikov2016seed,huang2018weakly}. After that, {step-3} uses the generated pseudo-masks as ground-truth annotations and combines the input image to train a fully-supervised semantic segmentation model. In the existing methods, for a fair comparison, DeepLab-v2~\cite{chen2017deeplab} is usually used as the fully-supervised model for predictions. 
However, despite the success of the existing WSSS methods, we can summarize that the multi-stage WSSS methods have problems of expensive computation costs (\eg, multiple networks are respectively needed in different training steps) and elaborate training procedures~\cite{araslanov2020single}. Even for some progressive methods, multiple human interventions are required~\cite{huang2018weakly}. For example, it is necessary to manually set thresholds from classification weights to select discriminative seed regions and design seed expansion methods. More importantly, each step in the multi-stage WSSS methods is executed separately. Especially for {step-1} and {step-2}, they are essentially disconnected from the third semantic segmentation model we ultimately want to obtain. This separation design does not intrinsically conform to the spirit of end-to-end training~\cite{fan2018end}.

To this end, a single-stage WSSS (SS-WSSS) model~\cite{araslanov2020single} was awakened, where inputs are images as well as the image-level class labels, and the outputs are pixel-level segmentation masks. SS-WSSS consists of a backbone network for generating CAMs, a normalized global weighted pooling module for generating preliminary segmentation masks, and a mask refinement module for result refinement. The whole model is end-to-end trained in a self-supervised manner. Although SS-WSSS simplifies the procedures of multi-stage WSSS models and can achieve somewhat satisfactory performance, as shown in the second column of Figure~\ref{fig1}, the results of such an immature model still suffer from the following two problems: (\uppercase\expandafter{\romannumeral1}) \textbf{\emph{background incompleteness}} and (\uppercase\expandafter{\romannumeral2}) \textbf{\emph{object incompleteness}}. For example, some background regions close to the ``\emph{person}'' (\ie, row (\uppercase\expandafter{\romannumeral1}) of Figure~\ref{fig1}) are regarded as the body of the ``\emph{person}'' and some parts of the ``\emph{airplane}'' (\ie, row (\uppercase\expandafter{\romannumeral2}) of Figure~\ref{fig1}) are lost in the segmentation mask. We empirically find that \textbf{\emph{background incompleteness}} is due to the fact that the model is not able to capture sufficient object context so that the object boundaries cannot be located accurately. Besides, \textbf{\emph{object incompleteness}} is caused by insufficient spatial details in high-level feature maps. Although some low-level features have been employed in the SS-WSSS model, there is intrinsically a large semantic gap between the low-level and the high-level features~\cite{zhang2020feature,zhang2018exfuse}, resulting in the detailed spatial information contained in the low-level feature maps not playing its role well when two levels of feature maps are fused. 

In this work, we propose a single-stage \textbf{W}eakly-\textbf{S}upervised \textbf{F}eature \textbf{C}oupling \textbf{N}etwork (\textbf{WS-FCN}), which can capture the multi-scale context formed from adjacent spatial feature grids, and encode fine-grained low-level features into high-level ones. \textbf{WS-FCN} is based on the SS-WSSS framework~\cite{araslanov2020single} with image-level class labels which has made up for the shortcomings of the multi-stage methods and is a progressive version. Specifically, a Flexible Context Aggregation (FCA) module is first proposed to compactly capture the context in different granular spaces so that the problem of \textbf{\emph{background incompleteness}} can be somewhat alleviated. Compared with the existing context aggregation modules, FCA brings a small amount of computational overhead on the basis of compact sampling (cf. Section~\ref{sec4:3}) while maintaining a higher performance. Besides, another learnable Semantically consistent Feature Fusion (SF2) module is proposed from the bottom (\ie, the low-level feature maps) to up (\ie, the high-level feature maps) to address the problem of \textbf{\emph{object incompleteness}}. SF2 projects feature maps of two levels with different semantic/contextual features into the same embedding space for fusion. Therefore, the problem of the semantic gap can be alleviated (cf. Section~\ref{sec4:4}).
To demonstrate the effectiveness of \textbf{WS-FCN}, we conducted extensive experiments on the challenging PASCAL VOC 2012~\cite{everingham2010pascal} and MS COCO 2014~\cite{lin2014microsoft} datasets. Experimental results declare that our \textbf{WS-FCN} can achieve new state-of-the-art results by $65.02\%$ and $64.22\%$ mIoU on the PASCAL VOC 2012 \emph{val} set and \emph{test} set, $34.12\%$ mIoU on the MS COCO 2014 \emph{val} set, respectively. Compared to the existing approaches, our \textbf{WS-FCN} is also more efficient.

The main contributions are summarized as follows: 1) we proposed a FCA module, which can capture the object context in multi-scale granular spaces. 2) we proposed a parameter-learnable SF2 module, which encodes the fine-grained low-level features into the high-level ones. 3) \textbf{WS-FCN} achieves the state-of-the-art results on both PASCAL VOC 2012 and MS COCO 2014 datasets. 

\section{Related Work}
\subsection{Semantic Segmentation}
Semantic segmentation has been studied tremendously in the past few years~\cite{chen2017deeplab,zhang2018exfuse,chen2018encoder,zhang2020causal,zhang2020feature,zeng2019joint,zhao2017pyramid,zhang2018context,zheng2021hierarchical,zhang2022graph}, where the fully convolutional network~\cite{long2015fully} is the most representative one and the majority of the existing excellent methods are based on this framework~\cite{zeng2019joint,zhang2020feature,zhang2018context,zhang2019co,yu2020context}. For semantic segmentation, one of the most important things is to guarantee the output feature maps have sufficient semantic and contextual information while ensuring a high resolution~\cite{zhang2023augmented}. To this end, the current methods can be divided into those aimed at maintaining high-resolution~\cite{chen2017rethinking,wang2020deep,yu2015multi} and those aimed at capturing the context~\cite{zhang2018context,zhang2019co,yu2020context,chen2018encoder}. To maintain a high resolution, the commonly used method is to fuse low-level feature maps into high-level feature maps before output~\cite{chen2018encoder,wang2020deep}, such that the output feature maps can meet the requirements of the high resolution. However, because feature maps of different levels usually have semantic gaps, \ie, high-level feature maps have more semantic information and vice versa, the effect of feature summation/concatenation is not satisfactory~\cite{zhang2020feature}. In this paper, we propose a parameter-learnable and semantically consistent feature fusion module from the low-level features to the high-level features. The proposed module projects feature maps with different granular information and semantic information into the same embedding space and then fuses them. Therefore, the problem of the semantic gap can be essentially alleviated.

\subsection{Weakly-Supervised Semantic Segmentation (WSSS)}
WSSS approaches are proposed to alleviate the problem of expensive labeling costs in fully-supervised settings. For example, it will cost more than $90$ minutes to label a pixel-level semantic mask for a $500 \times 500$ natural scene image~\cite{liu2021harmonic,yan2020social,zhang2020causal}. In particular, when the image to be labeled has a large size (\eg, images in cityscapes~\cite{cordts2016cityscapes} have the spacial size of $1024 \times 2048$), the time required for manual labeling is usually unbearable. To this end, WSSS methods have become the primary strategy to solve the problem of expensive labels. Most of the existing WSSS methods are multi-stage~\cite{ahn2018learning,ahn2019weakly,zhang2020causal,hong2017weakly,huang2018weakly,kolesnikov2016seed,zhang2018spftn,liu2020weakly,zhang2020splitting,chang2020weakly,chen2020weakly,feng2021deep,wang2022looking}. Among these methods, the image-level class labels have been widely used as weak supervision~\cite{ahn2019weakly,ahn2018learning,zhang2020causal,kolesnikov2016seed,liu2020weakly} because of their easy accessibility. For example, it only takes about one second to get the image-level class labels of an image. The main step for segmentation through this label is to obtain CAMs of training images. Then, based on CAMs, we can further obtain pseudo-masks of training images. However, CAMs can only localize the rough object regions in the image, so the obtained pseudo-masks are inaccurate. To alleviate this problem, some methods use additional labels or data (\eg, saliency~\cite{mehrani2010saliency,zhang2020splitting}, object proposals~\cite{wei2018ts2c}, and videos~\cite{hong2017weakly}) to provide more supervision during the training process. In this paper, we only use the image-level class labels as the supervision and the model is single-stage~\cite{araslanov2020single}. Besides, two schemes are proposed to alleviate problems of \textbf{\emph{background incompleteness}} and \textbf{\emph{object incompleteness}}.

\subsection{Context Aggregation}
Contextual information has empirically shown its effectiveness in a wide range of computer vision tasks, \eg, image classification~\cite{wang2018non,zhang2018recursive,chen20182}, image semantic segmentation~\cite{zhang2018context,zhang2019co,yang2021context,yu2020context}, instance segmentation~\cite{bolya2019yolact,chen2019hybrid}, object detection~\cite{zhang2020feature,wang2020deep}, and person re-identification~\cite{yan2018participation,TangLPT20}.  
The existing methods can be mainly divided into the following two types: multi-scale context aggregation from spatial adjacent pixels, \eg, ASPP~\cite{chen2017deeplab}, PPM~\cite{zhao2017pyramid} and MPM~\cite{hou2020strip}, and modeling the long-range dependencies, \eg, non-local interaction~\cite{wang2018non,zhang2020feature}, self-attention~\cite{vaswani2017attention,zhang2019co}, and object context~\cite{yuan2019object,zhang2019positional}, where every feature position participates in the global context calculation process. However, the existing context learning methods in semantic segmentation have the problem of either introducing unnecessary feature regions (especially for some slender objects) or excessive computation overhead~\cite{TANG2022108792,yan2020higcin}. For example, time and spatial costs for the non-local interaction and self-attention (single-head) are both $HW\times HW$, where $H$ and $W$ denote the width and height of the input feature maps, respectively. In our work, we also focus on the topic of context aggregation. Comparing to the previous ones, our proposed flexible context aggregation module is used to compactly capture the context in different granular spaces, which is a more effective context aggregation mechanism.
\section{Methodology}
In this section, we introduce the single-stage Weakly-Supervised Feature Coupling Network (WS-FCN) in detail. Specifically, we first give the problem statement of WSSS (in Section~\ref{sec3:1}). After that, implementation details of flexible context aggregation module (in Section~\ref{sec3:3}) and semantically consistent feature fusion module (in Section~\ref{sec3:4}) are introduced, respectively. Finally, we show the overall architecture of our proposed WS-FCN (in Section~\ref{sec3:2}).
\subsection{Problem Statement}
\label{sec3:1}
For a given image~${\bm I} \in \mathbb{R}^{3\times H\times W}$~(where~$H$ and~$W$ are the height and width, $3$ denotes the channel size) and the image-level class label~${\bm Y} \in \mathbb{R}^{C\times 1}$ (where~$C$ denotes class size), WSSS aims to make predictions for each pixel of the input image. To achieve this goal, most of the existing WSSS methods~\cite{ahn2018learning,zhang2020splitting,ahn2019weakly} obtain pseudo-masks via the multi-stage training procedures, which suffer from problems of the complicated training processes or time-consuming computations, or both. Besides, some human interventions are usually required~\cite{zhang2020causal,liu2020weakly}. To this end, the single-stage WSSS model~\cite{araslanov2020single} has been awakened via the self-supervised training manner in an end-to-end fashion. In this work, to alleviate problems of \textbf{\emph{background incompleteness}} and \textbf{\emph{object incompleteness}} in the single-stage WSSS~\cite{araslanov2020single}, we propose an effective yet efficient \textbf{WS-FCN}, which can mitigate the \textbf{\emph{background incompleteness}} problem by capturing multi-scale object context and address the \textbf{\emph{object incompleteness}} problem by fusing the fine-grained low-level features into the high-level features.
\begin{figure*}[t]
\centering
\includegraphics[width=0.95\textwidth]{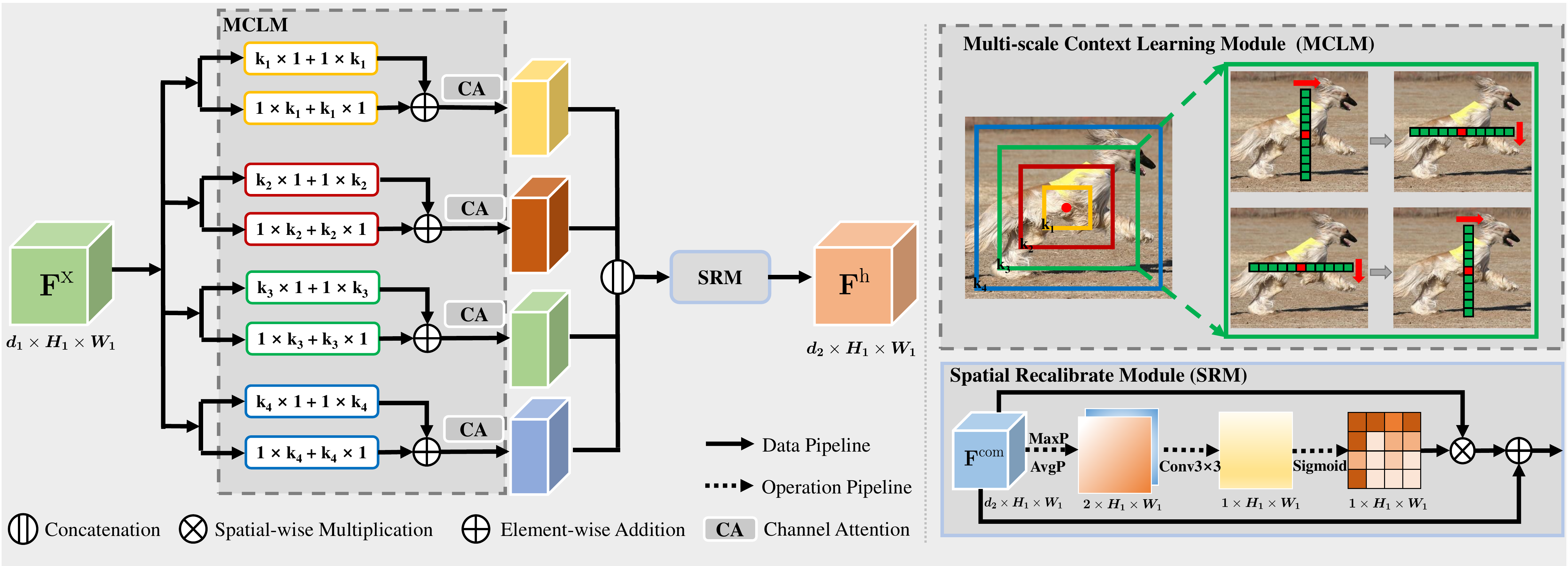}
\vspace{-2mm}
\caption{An illustration of the proposed Flexible Context Aggregation (FCA), which consists of the Multi-scale Context Learning Module (MCLM) and the Spatial Recalibration Module (SRM). In the upper right, the red point denotes an arbitrary feature position. Each square of different colors denotes the corresponding operation of four branches in the left side, and each operation involves the four strip shape convolutions.}
\label{fig3}
\vspace{-2mm}
\end{figure*}

\subsection{Flexible Context Aggregation~(FCA)}\label{sec3:3}
FCA is proposed to aggregate the global object context for mitigating the \textbf{\emph{background incompleteness}} problem. As illustrated in Figure~\ref{fig3}, FCA mainly consists of two modules: the Multi-scale Context Learning Module (MCLM) and the Spatial Recalibration Module (SRM). 

\myparagraph{MCLM.} For a set of feature maps ${\bm F}^{\mathrm x}\in {\mathbb R}^{d_{1}\times H_{1}\times W_{1}}$, which are generated by a fully convolutional backbone network~\cite{long2015fully}, we utilize different convolutional kernels on them to mine the multi-scale global contextual information. MCLM achieves it's effect in a combination of parallel row-wise and column-wise convolutions~($1\times k$ + $k\times 1$ and~$k\times 1$ + $1\times k$) manner. Instead of directly using the large kernel~$k\times k$ containing parameters of~$k^2$, MCLM can densely connect with the surrounding pixels in a~$k\times k$ region with parameters of~$4k$.
Concretely, we use parallel convolutions in each branch. In one branch, we perform convolution with the kernel size of~$k_{i}\times 1$ in a column-wise way followed by a kernel size of~$1\times k_{i}$ in a row-wise way. 
In the other branch, we perform convolution with a kernel size of~$1\times k_{i}$ in a row-wise way followed by a kernel size of~$k_{i}\times 1$ in a column-wise way. Here,~$i=1,2,3,4$ is used to distinguish four branches with different kernel sizes. Then, features containing the rich contextual information ${\bm F}^{k_{i}} \in {\mathbb R}^{d_{2}\times H_{1}\times W_{1}}$ in each branch are obtained. After that, a channel-wise attention~\cite{woo2018cbam} is utilized to increase the weights of object-related feature channels as well as suppress the non-significant feature channels. The above procedures can be formulated as:
\begin{gather}
{\bm F}^{k_{i}}={\bm f}^{k_{i}\times 1}({\bm f}^{1\times k_{i}}({\bm {F}^{\mathrm{x}}})) + {\bm f}^{1\times k_{i}}({\bm f}^{k_{i}\times 1}({\bm {F}^{\mathrm{x}}})), \label{1} \\  
\bm{M}^{i}_{c}= \bm\sigma ({\bm f}^{1\times 1}(\mathrm{GAP_{s}}(\bm{F}^{k_{i}})) +
                {\bm f}^{1\times 1}(\mathrm{GMP_{s}}(\bm{F}^{k_{i}}))),\label{2} \\ 
\bm{F}^{{k_{i}}'}=\bm{M}^{i}_{c}\circledast  \bm{F}^{k_{i}},
\label{3} 
\end{gather} 
where~${\bm f}^{k_{i}\times 1}$ and~${\bm f}^{1\times k_{i}}$ represent convolution layers with a kernel size of~${k_{i}\times 1}$ and~${1\times k_{i}}$ respectively. ~${\bm f}^{1\times 1}$ denotes a ${1\times 1}$ convolutional layer. $\bm \sigma(\cdot)$ is the sigmoid function.~$\mathrm{GAP_{s}(\cdot)}$ and~$\mathrm{GMP_{s}(\cdot)}$ denote global average pooling and global max pooling~\cite{he2016deep}. $\circledast $ denotes channel-wise multiplication, which is an element-wise multiplication along the channel dimension. Finally, we concatenate the feature maps of four branches to obtain the multi-scale feature maps~$\bm{F}^\mathrm{com} \in {\mathbb R}^{d_{2}\times H_{1}\times W_{1}}$ with a feature embedding as follows:
\begin{equation}
\begin{aligned}
\bm{F}^\mathrm{com}=\mathrm{E}(\mathrm{Cat}( \bm{F}^{k_{1}'},\bm{F}^{k_{2}'},\bm{F}^{k_{3}'},\bm{F}^{k_{4}'})), 
\label{4}
\end{aligned}
\end{equation}
where $\mathrm{Cat}(\cdot)$ denotes the concatenation operation along the channel dimension.~$\mathrm{E}(\cdot)$ denotes feature embedding operation, which consists of a $3\times3$ convolution, a BatchNorm layer~\cite{zhang2020causal}, and a ReLu layer~\cite{he2016deep}.
\begin{figure*}[tb]
\centering
\includegraphics[width=.95\textwidth]{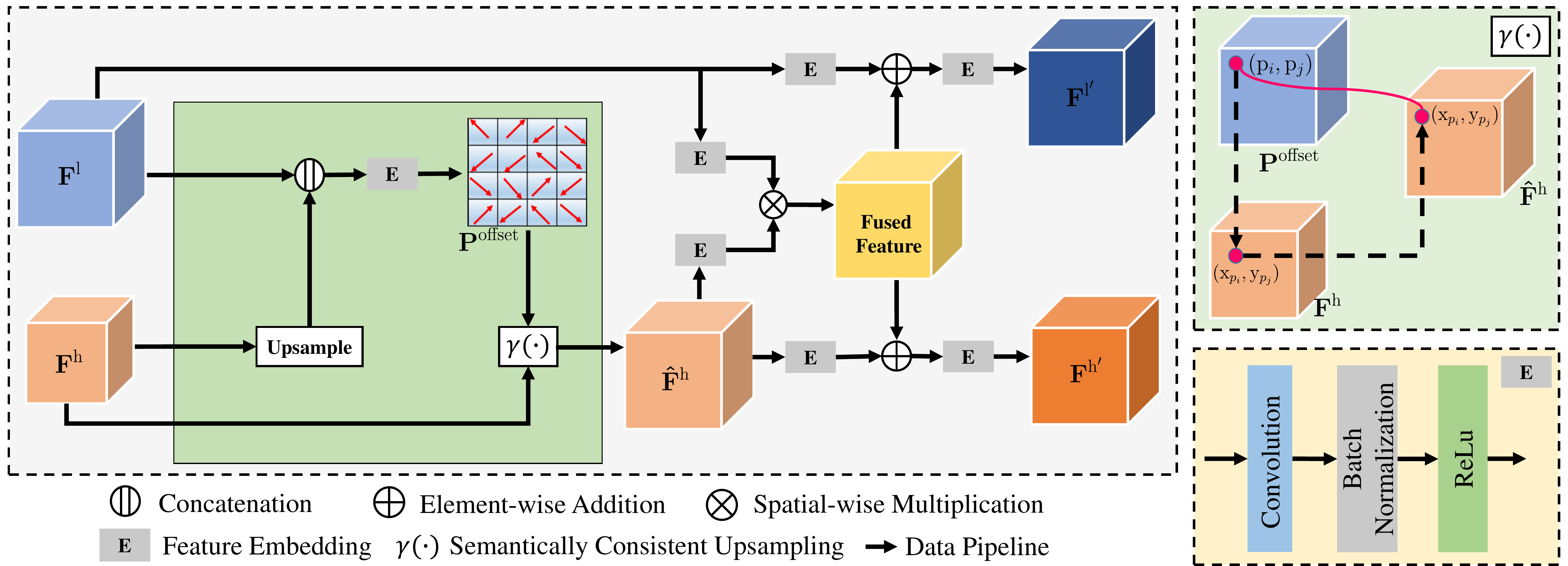}
\vspace{-2mm}
\caption{An illustration of the proposed SF2. The semantically aligned upsampling is proposed to upsample the low-resolution features as well as keeping the deep and shallow features aligned in semantics. $(p_i,p_j)$ represents the position corresponding to~${\bm{F}^{\mathrm h}}$.}
\vspace{-2mm}
\label{fig4}
\end{figure*}

\myparagraph{SRM.} SRM is proposed to enable the model to focus on the object regions, which can be expressed as:
\begin{equation}
\bm{M}_{s} =\bm\sigma ({\bm f}^{3\times 3}(\mathrm{Cat}(\mathrm{GAP_{c}}({\bm F}^\mathrm{com}),\mathrm{GMP_{c}}({\bm F}^\mathrm{com})))), 
\label{5}
\end{equation}

\begin{equation}
\bm{F}^\mathrm{h}=\bm{M}_{s}\otimes \bm{F}^\mathrm{com}+\bm{F}^\mathrm{com}, 
\label{6} 
\end{equation}
where~$\mathrm{GAP_{c}(\cdot)}$ and~$\mathrm{GMP_{c}(\cdot)}$ denote global average pooling and global max pooling along the channel dimension. $\otimes$ means spatial-wise multiplication, which is an element-wise multiplication along the spatial dimension.
We multiply the spatial attention map~$\bm{M}_{s}$ with~$\bm{F}^{\mathrm{com}}$ followed by summarizing~$\bm{F}^{\mathrm{com}}$. Compared to the previous context aggregation methods~\cite{yu2015multi,zhao2017pyramid,chen2017rethinking}, FCA can boost the final segmentation masks by: 1) capturing rich context information without much computational cost, and 2) preventing irrelevant regions from interfering with the prediction of segmentation masks.

\subsection{Semantically consistent Feature Fusion (SF2)}
\label{sec3:4}
To make full use both of the deep and shallow features to address the \textbf{\emph{object incompleteness}} problem, as shown in Figure~\ref{fig4}, we proposed a SF2 module.
Supposed high-level features~${\bm F}^{\mathrm h}\in {\mathbb R}^{d_{2}\times H_{1}\times W_{1}}$ and low-level features~${\bm F}^{\mathrm l}\in {\mathbb R}^{d_{2}\times H_{2}\times W_{2}}$ from the third backbone stage, {we first adopt the element-wise multiplication to fuse them.} Then, the fused features add with~${\bm F}^{\mathrm l}$ and~${\bm F}^{\mathrm h}$ to obtain refined shallow feature maps~${\bm F}^{\mathrm l'}$ and deep feature maps~${\bm F}^{\mathrm h'}$, which can propagate the advantages of low/high-level feature maps to each other. 
Compared to other fusion operations~\cite{zhang2020feature,zhao2017pyramid,zhang2018context,Wang_2022_IJCAI}, SF2 can neither deteriorate the original feature representations nor make refined feature maps focus on irrelevant regions.
The above procedures can be formulated as:
\begin{equation}
\bm{F}^{\mathrm h'}=\mathrm{E}(\mathrm{E}(\bm\gamma (\bm{F}^{\mathrm h}))+\mathrm{E}(\bm\gamma (\bm{F}^{\mathrm h}))\otimes  \mathrm{E}(\bm{F}^{\mathrm l})), \label{7} 
\end{equation}

\begin{equation}
\bm{F}^{\mathrm l'}=\mathrm{E}(\mathrm{E}(\bm{F}^{\mathrm l})+\mathrm{E}(\bm\gamma({\bm{F}}^{\mathrm h}))\otimes \mathrm{E}(\bm{F}^{\mathrm l})), \label{8} 
\end{equation}
where~$\bm\gamma(\cdot)$ denotes the general semantically consistent upsampling scheme. $\otimes$ denotes the element-wise multiplication.

To be specifical, the classical upsampling is executed to keep deep and shallow feature maps the same in spatial resolution in Eq.(\ref{7}) and Eq.(\ref{8}) before feature fusion. Given the low-resolution feature map~${\bm F}^{\mathrm h}\in {\mathbb R}^{d_{2}\times H_{1}\times W_{1}}$, we can obtain the enlarged high-resolution feature map~${\hat{\bm F}}^{\mathrm h}\in {\mathbb R}^{d_{2}\times H_{2}\times W_{2}}$ by bilinear upsampling as follows:
\begin{gather}
\hat{\bm F}^{\mathrm h}_{\bm{p}^{\mathrm h}}=\bm\gamma (\bm{F}^{\mathrm h}) =\sum_{\bm{p}\in {\bm N}(\bm{p}^{\mathrm l})} \bm{W_{p}}\bm{F}^{\mathrm h}_{\bm{p}}, \label{9} \\
\bm{p}^{\mathrm l}=\mathcal{\bm R}(\bm{p}^{\mathrm h})=\frac{\bm{p}^{\mathrm h}}{\bm{s}},\label{10} 
\end{gather}
where~$\bm{p}^{\mathrm h}$ and~$\bm{p}^{\mathrm l}$ represent the position on the spatial grid of~$\hat{\bm F}^{\mathrm h}$ and~$\bm{F}^{\mathrm h}$, respectively.~$\bm{s}$ means upsampling stride of~$\bm{F}^{\mathrm h}$.~$\bm{N}(\bm{p}^{\mathrm l})$ indicates the neighbors of~${\bm p}^{\mathrm l}$ in~$\bm{F}^{\mathrm h}$.~$\bm{W_{p}}$ denotes the weights measured by the distance to the~${\bm p}^{\mathrm l}$. In general, from low-resolution to high-resolution transformation, we can attribute to learning a mapping relation~$\mathcal{\bm R}:\bm{p}^{\mathrm h}\rightarrow{\bm p}^{\mathrm l}$ to decide the position correspondence between high-resolution and low-resolution feature maps. 
Although bilinear upsampling learns the mapping relation~$\mathcal{\bm R}$ via a uniform sampling strategy, there exists semantic information loss between deep and shallow feature maps due to downsampling. It means~$\hat{\bm F}^{\mathrm h}$ still has semantic deviation with the $\bm{F}^{\mathrm l}$ of the same spatial resolution, which has an influence on the following feature fusion. 

To mitigate this problem, we propose a novel feature fusion approach inspired by~\cite{li2020semantic}, which can dynamically track the semantic ``moving'' direction and adaptively adjust the sampling positions to align~$\hat{\bm F}^{\mathrm h}$ with~$\bm{F}^{\mathrm l}$ in semantics as shown in Figure~\ref{fig4}. Concretely, we first learn a semantic offset~$\bm{p}^{\textup{offset}}\in {\mathbb R}^{2\times H_{2}\times W_{2}}$ of each pixel in~$\hat{\bm F}^{\mathrm h}$ by concatenating~$\bm{F}^{\mathrm l}$ and upsampled~${\bm F}^{\mathrm h}$. Then,~$\bm{p}^{\textup{offset}}$ adds with~$\bm{p}^{\mathrm h}$ so that SF2 can adaptively adjust the sampling positions on~$\bm{F}^{\mathrm h}$ according to~$\bm{p}^{\textup{offset}}$. Finally, a dynamic position correspondence between deep and shallow feature maps has been learned.  
\begin{equation}
\begin{aligned}
\bm{p}^{\mathrm{offset}}&=\mathrm{E}(\mathrm{Cat}(\mathrm{Up}(\bm{F}^{\mathrm h}),\bm{F}^{\mathrm l})), \label{11}
\end{aligned}
\end{equation}

\begin{equation}
\begin{aligned}
\bm{p}^{\mathrm l}&=\mathcal{\bm R}(\bm{p}^{\mathrm h})=\frac{\bm{p}^{h}+\bm{p}^{\mathrm{offset}}}{\bm{s}},\label{12}
\end{aligned}
\end{equation}
where~$\mathrm{Up}(\cdot)$ denotes the bilinear upsampling. By the semantically aligned upsampling approach, the semantic misalignment problem of different-level feature maps can be alleviated, which is beneficial for feature fusion. In general, the proposed SF2 module improves the segmentation performance by making shallow and deep feature maps semantic alignment in the same spatial resolution first and fusing the fined-grained shallow feature maps into deep feature maps then. 

\subsection{Overview}\label{sec3:2}
\begin{figure*}[!t]
\centering
\includegraphics[width=.99\textwidth]{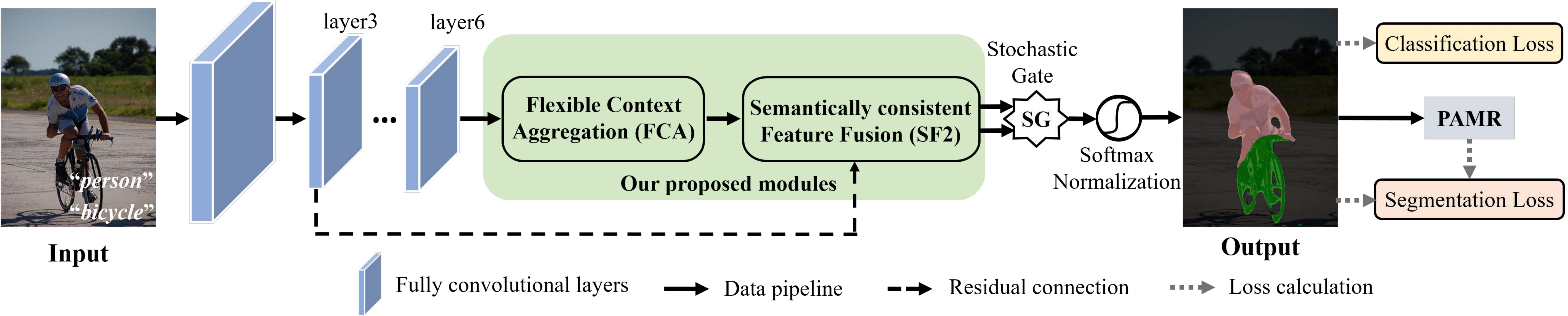} 
\caption{Overview of our proposed \textbf{WS-FCN}. A single network is used to achieve the weakly-supervised semantic segmentation objective in an end-to-end training fashion, which comprehensively mines the global context and the local contents via FCA and SF2, respectively.} 
\label{overview}
\end{figure*}

Figure~\ref{overview} gives an overall of our proposed \textbf{WS-FCN}, which mainly consists of five components: a fully convolutional network served as the backbone, a FCA module, a SF2 module, a stochastic gate and a Pixel-Adaptive Mask Refinement (PAMR) module~\cite{araslanov2020single}. First, an input image with the image-level class labels are sent into the backbone network which is pre-trained on the ImageNet~\cite{deng2009imagenet} for feature extraction. Then, the extracted features from the last stage of the backbone which are $1/8$ spatial size of the input image are sent into the FCA module, which is used to capture the multi-scale context information via the multiple strip convolutions in different granular spaces. After that, we use the proposed SF2 module to fuse features from both the third backbone stage and outputs of the FCA module in a semantically consistent manner. Then, we adopt a stochastic gate and a softmax normalization layer to improve the model generalization ability and regularize feature maps, respectively. After that, feature maps and image-level class labels generated at this stage are used to calculate a classification loss. Finally, the PAMR module is implemented to refine the segmentation mask. The segmentation loss is implemented before and after PAMR.
\section{Experiments}
\subsection{Datasets and Evaluation Metrics}\label{sec4:1}
\myparagraph{Datasets.}
In our work, the commonly used and challenging benchmarks PASCAL VOC 2012~\cite{everingham2010pascal} and MS COCO 2014~\cite{lin2014microsoft} are used for weakly supervised semantic segmentation. PASCAL VOC 2012 has $20$ classes of objects in total $4,369$ images, where the \emph{training} set, the \emph{val} set and the \emph{test} set contains $1,464$ images, $1,449$ images and $1,456$ images, respectively. Particularly, following~\cite{zhang2020causal,wei2018revisiting}, the augmented \emph{training} set is used for a fair experimental comparison which has $10,582$ training images. MS COCO 2014 contains $81$ classes of objects (including one background), and it has $80$K images for \emph{training} and $40$K images for \emph{val}.

\myparagraph{Evaluation metrics.}
Following~\cite{chen2017deeplab,zhong2020squeeze}, the standard Pixel Accuracy (PixAcc) and mean Intersection of Union (mIoU) are used as the evaluation metrics on model effectiveness. Besides, to verify the model efficiency, the model Parameters (Params) is also taken into account.

\subsection{Backbone and Baseline}\label{sec4:2}
\myparagraph{Backbone.} In this paper, we choose the commonly used WideResNet-38~\cite{wu2019wider} as the backbone network for feature extraction, which is used in ablation studies and performance comparisons with state-of-the-art methods. In particular, to ensure a high-resolution output scale for feature maps and not have too much parameter growth, we replace the traditional convolutions of the last two blocks with dilated/atrous convolutions as in~\cite{ahn2018learning}, such that the spatial scale of output feature maps is $1/8$ of the input image. All our implementations are deployed on the PyTorch~\cite{paszke2019pytorch} deep learning platform.

\myparagraph{Baseline.}
Our baseline is the single-stage semantic segmentation~\cite{araslanov2020single} model, which is based on the self-supervised training manner with only image-level class labels. In the training stage, the backbone network first generates the initial CAMs. Then, a mask refinement module named PAMR is used to refine CAMs as pseudo ground-truth to complete the self-supervised training process. In this work, our proposed \textbf{WS-FCN} also follows the same training procedure. 

\myparagraph{Settings.}
As in~\cite{araslanov2020single}, the whole network is first pre-trained on ImageNet~\cite{deng2009imagenet}, and then fine-tuned on the augmented \emph{training} set of PASCAL VOC 2012~\cite{everingham2010pascal} and \emph{training} set of MS COCO 2014~\cite{lin2014microsoft} on two GeForce GTX 1080 Ti GPUs, respectively. The SGD strategy is used as the optimizer, where the weight decay and momentum are set to $5\times 10^{-4}$ and $0.9$, respectively. The base constant learning rate is set to $0.001$ for the whole model. Peculiarly, in order to speed up the model training process and accelerate the convergence process, the learning rate of our proposed modules (\ie, FCA and SF2) is set $20\times$ (\ie, $0.001$) of the base learning rate as in~\cite{zhang2018context,zhang2020causal}.
In FCA, in order to avoid introducing too many model parameters, we set the kernel size of $k = [1, 3, 5, 7]$. The whole model is trained for $30$ epochs. Concretely, we first train our model for $7$ epochs using only the multi-label classification loss (\ie, training the multi-label image classification model), and then switch on the self-supervised semantic segmentation loss (\ie, training the semantic segmentation model) for the remaining $23$ epochs. For data augmentation, following~\cite{ahn2018learning,wei2018revisiting}, we first use horizontal flipping and random resizing under the range of $0.9$ to $1.0$. Then, the training image is randomly cropped into the fixed size of $321\times 321$.

\begin{table}[t] 
\begin{center}
\renewcommand\arraystretch{1.4}
\setlength{\tabcolsep}{.5pt}{
\caption{Ablation studies on the \emph{val} set of PASCAL VOC 2012~\cite{everingham2010pascal}. ``Baseline'' result is the our re-implemented one and the code is obtained from its GitHub page with nothing changes. ``FCA$_{\textrm{MCLM}}$'' and ``FCA$_{\textrm{SRM}}$'' denotes that implementing MCLM, and SRM on FCA, respectively.}
\begin{tabular}{ c  c  c  c | c  c  c } 
Baseline~\cite{araslanov2020single} & FCA$_{\textrm{MCLM}}$ & FCA$_{\textrm{SRM}}$ & SF2 & mIoU (\%) & PixAcc (\%) & Params (M) \\ 
\hline \hline
\cmark & \xmark & \xmark & \xmark & 58.38 & 87.58 & 137.10 \\ 
\cdashline{1-7}[0.8pt/2pt]
\cmark & \cmark & \xmark & \xmark & 59.71$_{\color{red}{+1.33}}$ & 88.41$_{\color{red}{+0.83}}$ & 143.52 \\ 
\cmark & \cmark & \cmark & \xmark & 60.13$_{\color{red}{+1.75}}$ & 88.55$_{\color{red}{+0.97}}$  & 143.52 \\ 
\cmark & \xmark & \xmark & \cmark & 59.45$_{\color{red}{+1.07}}$ & 89.00$_{\color{red}{+1.42}}$  & 140.92 \\
\cmark & \cmark & \cmark & \cmark & 60.68$_{\color{red}{+2.30}}$ & 89.32$_{\color{red}{+1.74}}$  & 147.34 \\ 
\end{tabular}\label{tab1}}
\end{center}
\end{table}
\subsection{Ablation Study}\label{sec4:3}
Our ablation study aims to investigate the effectiveness of FCA and SF2 as well as their combination in segmentation. To this end, we conduct a series of experiments on the \emph{val} set of PASCAL VOC 2012~\cite{everingham2010pascal}. Table~\ref{tab1} shows the experimental results against the baseline. The experimental results are based on the single-scale testing strategy. The model parameters and pixel accuracy are listed for comparison. 

\myparagraph{Effectiveness of FCA.}
As shown in Table~\ref{tab1}, we can observe that with the help of the complete version of FCA (\ie, FCA$_{\textrm{ALL}}$), the model can achieve the mIoU by $60.13\%$ and PixAcc by $88.55\%$. Compared to the baseline model~\cite{araslanov2020single}, FCA has a performance gain of $1.75\%$ mIoU and $0.97\%$ PixAcc, respectively. 
Besides, we also provide a subset of FCA, \ie, only implementing MCLM on the baseline model (\ie, FCA$_{\textrm{MCLM}}$). Results show that MCLM boosts the model performance of $1.33\%$ mIoU and $0.83\%$ PixAcc, respectively. These results empirically illustrate that aggregating multi-scale context information as well as suppressing the useless channel information play an important role in WSSS.

\myparagraph{Effectiveness of SF2.} 
From the bottom part of Table~\ref{tab1}, compared to results of the baseline model~\cite{araslanov2020single}, we can observe that SF2 has the performance gain by mIoU of $1.07\%$ and PixAcc of $1.42\%$, which not only verifies the importance of merging shallow feature maps into deep ones for WSSS task, but also verifies the necessity of semantic information consistency for feature fusion of different layers. 

\myparagraph{Effectiveness of both FCA and SF2.}
When both FCA and SF2 are implemented on the baseline model, we can achieve the best performance of $60.68\%$ mIoU and $89.32\%$ PixAcc, which has a performance gain of $2.30\%$ mIoU and $1.74\%$ PixAcc on the baseline model, respectively. It demonstrates that these two proposed modules do not conflict in practice and can jointly boost the performance of the baseline model. 

\begin{figure*}[t]
\centering
\includegraphics[width=.99\textwidth]{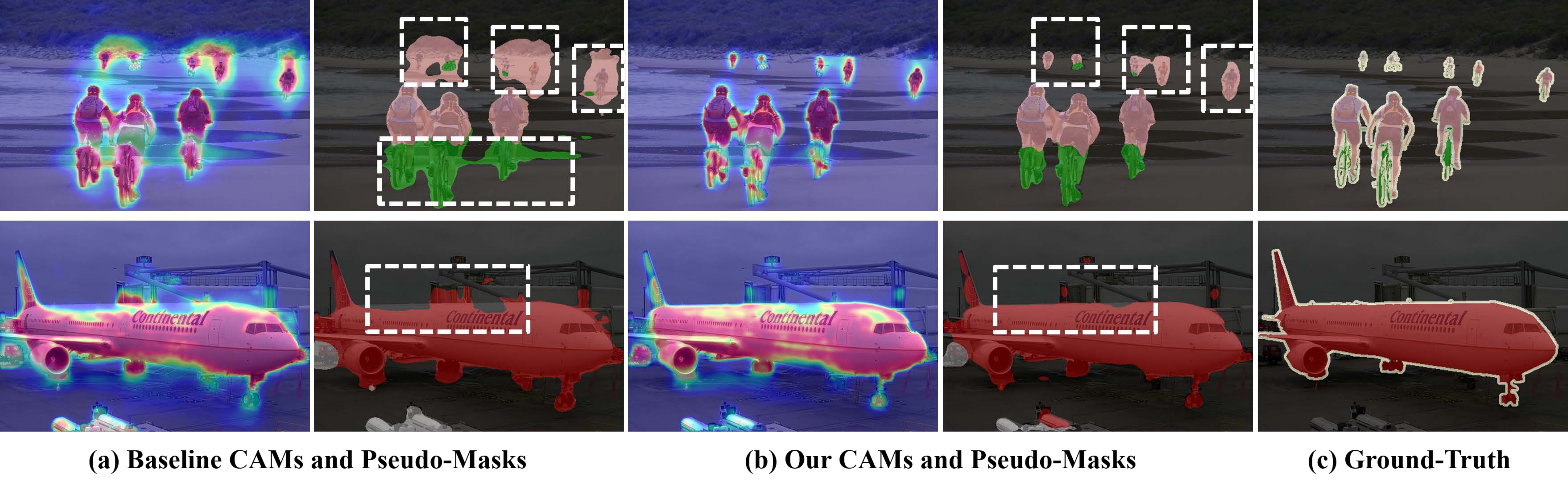} 
\caption{The visualizations of CAMs and segmentation masks. Compared to the baseline model, CAMs of our model can be localized more accurately and lead to higher quality of segmentation masks. The white dashed frames highlight the improved areas predicted by our WS-FCN. It is worth noting that the pixel-level ground-truth is used to emphasize the effectiveness of our method and is not used in the model training process.}
\label{fig5}
\end{figure*}

To validate the effectiveness of FCA and SF2, we give visualization comparisons generated by the baseline~\cite{araslanov2020single} and our \textbf{WS-FCN} in Figure~\ref{fig5}. We show some examples from the \emph{training} set of PASCAL VOC 2012~\cite{everingham2010pascal}. We can observe that CAMs generated by the baseline model have some over-activation regions, \eg, the ``\emph{bicycle}". In contrast, \textbf{WS-FCN} can alleviate this phenomenon and obtain accurate object boundaries. Especially, we can obviously observe that the results generated by \textbf{WS-FCN} can not only accurately cover the object area, but also do not produce the problem of mask redundancy. It indicates FCA can capture sufficient object context to locate object boundaries. Besides, when we deploy SF2 on the baseline model, the model can well introduce sufficient spatial details to deep features like the ``\emph{airplane}", which demonstrates our SF2 can complete the lacking regions.
\begin{table}[t] 
\begin{center}
\renewcommand\arraystretch{1.4}
\setlength{\tabcolsep}{2pt}{
\caption{Superiority of FCA and SF2 on the \emph{val} set of PASCAL VOC 2012~\cite{everingham2010pascal}. ``+'' denotes deploying the corresponding scheme on the baseline model.``Params" here means the parameters of the whole network.}
\begin{tabular}{ l c  c  c | c  c  c } 
Settings &   &   &   & mIoU (\%) & PixAcc (\%) & Params (M)  \\ 
\hline \hline
baseline~\cite{araslanov2020single} &    &    &   &58.38 & 87.58 &  137.10 \\
\cdashline{1-7}[0.8pt/2pt]
+ Non-local~\cite{wang2018non} &    &    &   &53.82  & 86.12 & 140.97 \\ 
+ Large Conv~\cite{lecun-nature-15} ($7 \times 7$) &    &    &   & 54.85 & 86.95 & 157.75 \\ 
+ PPM~\cite{zhao2017pyramid}  &   &    &   & 56.32 & 87.30 &  142.03 \\   
+ \textbf{FCA} (ours) &    &    &   & \textbf{60.13} & \textbf{88.55} & \textbf{143.52} \\   
\cdashline{1-7}[0.8pt/2pt]
+ Summation  &    &    &   & 56.45 & 87.78 & 140.85 \\ 
+ Concatenation  &   &    &   & 55.20 & 87.10 & 142.03 \\  
+ Multiplication  &  &    &   & 57.11 & 87.91 & 140.85 \\  
+ \textbf{SF2} (ours) &    &    &   & \textbf{59.45} & \textbf{89.00} & \textbf{140.92}
\end{tabular}\label{tab2}}
\vspace{-2mm}
\end{center}
\end{table}
\subsection{Superiority of FCA and SF2}\label{sec4:4}
Our proposed FCA and SF2 are used for the global context aggregation and feature fusion, respectively. In this section, to further demonstrate the superiority of our proposed FCA and SF2 modules, we compare them against other prevalent context aggregation~\cite{wang2018non,zhao2017pyramid,chen2017rethinking} and commonly used feature fusion methods. 

\myparagraph{Superiority of FCA.}
We validate the superiority of FCA by comparing it with the non-local operation~\cite{wang2018non}, large conv~\cite{lecun-nature-15} operation, pyramid pooling module (PPM)~\cite{zhao2017pyramid} and atrous spatial pyramid pooling (ASPP) operation~\cite{chen2017rethinking} (has been used in the baseline model). As shown in Table~\ref{tab2}, we can observe that with the help of FCA, our model achieves a significantly higher performance of $60.13\%$ mIoU and $88.55\%$ PixAcc, which surpasses these five context aggregation methods by a large margin. Comparatively, the progressive non-local operation~\cite{wang2018non} only obtains $53.82\%$ mIoU and $86.12\%$ PixAcc. The baseline model has better results than the non-local operation~\cite{wang2018non} and the large conv operation~\cite{lecun-nature-15}, which can achieve $58.38\%$ on mIoU and $87.58\%$ on PixAcc, respectively. Although the baseline model uses the ASPP operation~\cite{chen2017rethinking} to capture the multi-scale context and the segmentation performance is improved, there are largely semantic and contextual gaps between different scale feature maps. Comparatively, our proposed FCA not only extracts multi-scale features, but also self-corrects these features by the channel attention manner.
From the perspective of model parameters, the baseline model uses the least model parameters by $137.10$ M, while our FCA has more computation parameters and slightly surpasses it by $6.42$ M. This is because our FCA can non-destructively pick up four different kernel sizes and add attention mechanisms that can capture more global information and suppress useless channel information. However, compared to the performance increasement, we graciously argue this parameter increasement is worth it. The experimental results confirm that our FCA can somewhat alleviate the \textbf{\emph{background incompleteness}} problem without introducing too many model parameters. 
\begin{table}[t] 
\begin{center}
\renewcommand\arraystretch{1.4}
\setlength{\tabcolsep}{5pt}{
\caption{More experimental results using multi-scale and flip operations in the inference stage.}
\begin{tabular}{ l  c  c | c  c } 
Settings & Multi-Scale &  Flips & mIoU (\%) & PixAcc (\%) \\ 
\hline \hline
WideResNet-38 & \xmark & \xmark &59.92 & 89.14 \\    
\cdashline{1-5}[0.8pt/2pt]
WideResNet-38 & \cmark & \xmark & 61.13$_{\color{red}{+1.21}}$ & 89.41$_{\color{red}{+0.27}}$ \\ 
WideResNet-38 & \xmark & \cmark & 60.68$_{\color{red}{+0.76}}$ & 89.32$_{\color{red}{+0.18}}$ \\   
WideResNet-38 & \cmark & \cmark & 61.57$_{\color{red}{+1.65}}$ & 89.54$_{\color{red}{+0.40}}$ \\ 
\end{tabular}
\label{tab3}}
\end{center}
\vspace{-4mm}
\end{table}

\myparagraph{Superiority of SF2.}
We also explore the effectiveness of different feature fusion methods. As shown in Table~\ref{tab2}, we can observe that summation or concatenation of features can not significantly improve the model accuracy. The reason is that there is a big semantic gap between deep feature maps and shallow feature maps. Although this kind of feature fusion operations can be implemented through the unification of the feature scale, the problem of the semantic gap exists. Element-wise multiplication can achieve better performance by $57.11\%$ mIoU and $87.91\%$ PixAcc, while our SF2 can surpass it by $2.34\%$ mIoU and $1.09\%$ PixAcc, which verifies the importance of semantic alignment between deep and shallow feature maps. At this point, the problem of the semantic gap can be alleviated. In terms of the model efficiency, the summation and multiplication bring the parameters by $140.85$ M. And our SF2 is much more efficient than the feature concatenation operation which has reduced by $1.11$ M. In brief, SF2 can effectively align deep features with shallow features in semantics as well as fuse the spatial detail information into deep features to fill the \textbf{\emph{object incompleteness}}.

\subsection{More Experimental Analysis}\label{sec4:5}
Table~\ref{tab3} gives the performance contribution of using multi-scale and flip operations on the \emph{val} set of PASCAL VOC 2012 which shows the effectiveness of scaling images including~[1,~0.5,~1.5,~2] and horizontal flip. We can observe that both of them have a beneficial effect on improving the segmentation performance. Specifically, when we only adopt multi-scale operation, it can improve the model performance by $1.21\%$ mIoU and $0.27\%$ PixAcc, respectively. In addition, a solely flip operation can boost the performance by $0.76\%$ mIoU and $0.18\%$ PixAcc. When we adopt both of them during the inference stage, we can achieve $61.57\%$ mIoU and $89.54\%$ pixAcc. This shows that the continued addition of some inference tricks to our model can still bring sustained performance gains.
 
\subsection{Comparisons with the State-of-the-arts}\label{sec4:6}
In this section, we make result comparisons between our proposed \textbf{WS-FCN} and the state-of-the-art WSSS approaches quantitatively and qualitatively.

\myparagraph{Quantitative results on PASCAL VOC 2012~\cite{everingham2010pascal}}.
To demonstrate the effectiveness of the proposed \textbf{WS-FCN}, we compare it against the existing state-of-the-art weakly-supervised semantic segmentation approaches on the \emph{val} and \emph{test} set of PASCAL VOC 2012. To make fair comparisons, we adopt multi-scale and flip strategies in the inference stage. As shown in Table~\ref{tab4}, when compared to the weakly supervised and single stage methods~\cite{papandreou2015weakly,pinheiro2015image,roy2017combining,araslanov2020single} with only image-level class labels as the supervision, we can observe that our approach can achieve the best performance by $63.23\%$ mIoU on the \emph{val} set and $64.22\%$ mIoU on the \emph{test} set\footnote{http://host.robots.ox.ac.uk:8080/anonymous/QYHIA1.html} with the help of the CRF. Specifically, compared to the recent SSSS~\cite{araslanov2020single}, our \textbf{WS-FCN} without the CRF operation outperforms it with the same backbone by $1.87\%$ mIoU on the \emph{val} set and $1.80\%$ mIoU on the \emph{test} set, respectively. Besides, the performance of our approach with CRF is even close to the Joint Saliency~\cite{zeng2019joint} method, which uses DenseNet-169 as the backbone network and saliency as the extra supervisions. 
Compared to some multi-stage semantic segmentation methods with only image-level class labels, our method is a little inferior to the NSROM~\cite{yao2021non}, MCIS~\cite{wang2022looking}, RIB~\cite{lee2021reducing} and other advanced methods~\cite{ahn2019weakly,shimoda2019self,zhang2020causal}. However, compared to the state-of-the-art RIB~\cite{lee2021reducing} method, our method can still achieve $92.58\%$ of its performance on the \emph{val} set, which can demonstrate the effectiveness of our \textbf{WS-FCN}. Besides, it can be observed that the performance of our method with the CRF operation surpasses AffinityNet~\cite{ahn2018learning} by $1.53\%$ mIoU on \emph{val} set and $0.52\%$ mIoU on \emph{test} set. It means that our method has the superior capability to locate and mine the entire object region in a self-supervised training manner. Although there is still a gap between our method and some methods like EPS~\cite{lee2021railroad}, DRS~\cite{kim2021discriminative} and OAA+~\cite{jiang2021online}, our WS-FCN achieves better performance than DCSP~\cite{chaudhry2017discovering}, RDC~\cite{wei2018revisiting} and DSRG~\cite{huang2018weakly} methods, which validates the capacity of our method to boost the performance of segmentation masks.
We can further improve the model performance on the setting of removing the masks of false positive classes using ground-truth image-level class labels, in which we can achieve $64.02\%$ mIoU and $65.02\%$ mIoU with CRF on the \emph{val} set, respectively.
\begin{table}[t!]
\begin{center}
\renewcommand\arraystretch{1.6}
\setlength{\tabcolsep}{0.5pt}{
\caption{Comparisons with the state-of-the-art methods on the \emph{val} and \emph{test} sets of Pascal VOC 2012~\cite{everingham2010pascal} in terms of mIoU ($\%$). ``Superv." is the training supervision ($\mathcal{F}$: pixel-level mask, $\mathcal{I}$: image-level class label, $\mathcal{S}$: saliency masks, $\mathcal{D}$: extra data). ``--" denotes that there is no reported result in its paper. ``\S" represents the results that we use ground-truth of image-level class labels to remove masks of false positive classes predicted by our model. ``*" means that there is no ground-truth of image-level class labels in this setting.}
\begin{tabular}{ l c c | c c } 
Methods & Backbone & Superv. & \emph{val} (\%) & \emph{test} (\%) \\
\hline \hline
\multicolumn{4}{l}{(\textbf{a}) \textbf{\emph{Fully-supervised}}} \\ 
\cdashline{1-5}[0.8pt/2pt]
WideResNet-38~\cite{wu2019wider} & -- & $\mathcal{F}$  & 80.80  & 82.50 \\
DeepLab-v2~\cite{chen2017deeplab} & ResNet-101 & $\mathcal{F}$  & --  & 79.70 \\ 
DeepLab-v3+~\cite{chen2018encoder} & Xception-65~\cite{chollet2017xception} & $\mathcal{F}$  & --  & 87.80 \\ 
\hline \hline
\multicolumn{4}{l}{(\textbf{b}) \textbf{\emph{Multi-stage weakly-supervised + Saliency}}} \\ 
\cdashline{1-5}[0.8pt/2pt]
STC~\cite{wei2016stc} & DeepLab~\cite{chen2017deeplab} & $\mathcal{I,S,D}$   & 49.80  & 51.20 \\
Saliency~\cite{oh2017exploiting} & VGG-16 & $\mathcal{I,S}$  & 55.70  & 56.70 \\
DCSP~\cite{chaudhry2017discovering} & ResNet-101 & $\mathcal{I,S}$ & 60.80  & 61.90 \\
RDC~\cite{wei2018revisiting} & VGG-16 & $\mathcal{I,S}$ & 60.40  & 60.80 \\
DSRG~\cite{huang2018weakly} & ResNet-101 & $\mathcal{I,S}$ & 61.40  & 63.20 \\
FickleNet~\cite{lee2019ficklenet} & ResNet-101 & $\mathcal{S}$  & 64.90  & 65.30 \\
EPS~\cite{lee2021railroad} & VGG-16 &$\mathcal{I,S}$  & 66.60  & 67.90 \\
{DRS~\cite{kim2021discriminative}} &  {ResNet-101} & {$\mathcal{I,S}$} & {71.20}  & {71.40} \\
 {OAA+~\cite{jiang2021online}} &  {ResNet-101} & {$\mathcal{I,S}$} & {66.10}  & {67.20}\\
\hline \hline
\multicolumn{4}{l}{(\textbf{c}) \textbf{\emph{Multi-stage weakly-supervised}}} \\ 
\cdashline{1-5}[0.8pt/2pt]
SEC~\cite{kolesnikov2016seed} & VGG-16 & $\mathcal{I}$  & 50.70  & 51.70 \\
AffinityNet~\cite{ahn2018learning} & WideResNet-38 & $\mathcal{I}$  & 61.70  & 63.70 \\
IRN~\cite{ahn2019weakly} & ResNet-50 & $\mathcal{I}$  & 63.50   & 64.80 \\
SSDD~\cite{shimoda2019self} & WideResNet-38 &  $\mathcal{I}$  & 64.90  & 65.50  \\
CONTA~\cite{zhang2020causal} & ResNet-50 & $\mathcal{I}$  &  65.30   & 66.10 \\
NSROM~\cite{yao2021non} & VGG-16 & $\mathcal{I}$  & 65.50  & 65.30  \\
 {MCIS~\cite{wang2022looking}} &  {ResNet-101} &  {$\mathcal{I}$}  &  {66.20}  & {66.90}  \\
RIB~\cite{lee2021reducing} & ResNet-101 & $\mathcal{I}$  &  68.30 & 68.60\\
\hline \hline
\multicolumn{4}{l}{(\textbf{d}) \textbf{\emph{Single-stage + Saliency}}} \\ 
\cdashline{1-5}[0.8pt/2pt]
Joint Saliency~\cite{zeng2019joint} & DenseNet-169 & $\mathcal{S,D}$  & 63.30  & 64.30 \\
\hline \hline
\multicolumn{4}{l}{(\textbf{e}) \textbf{\emph{Single-stage}}} \\ 
\cdashline{1-5}[0.8pt/2pt]
TransferNet~\cite{hong2016learning} & VGG-16 &  $\mathcal{D}$  & 52.10  & 51.20 \\
WebCrawl~ \cite{hong2017weakly} & VGG-16 & $\mathcal{D}$  & 58.10  & 58.70 \\
EM~\cite{papandreou2015weakly} & VGG-16 & $\mathcal{I}$  & 38.20  & 39.60  \\
MIL-LSE~\cite{pinheiro2015image} & Overfeat~\cite{sermanet2014overfeat}  & $\mathcal{I}$ & 42.00  & 40.60 \\
CRF-RNN~\cite{roy2017combining} & VGG-16 & $\mathcal{I}$  & 52.80   & 53.70 \\
SSSS~\cite{araslanov2020single} & WideResNet-38 & $\mathcal{I}$  & 59.70  & 60.50 \\
SSSS~\cite{araslanov2020single} + CRF & WideResNet-38 & $\mathcal{I}$  & 62.70  & 64.30 \\
\cdashline{1-5}[0.8pt/2pt]
\textbf{WS-FCN} (ours) & WideResNet-38 & $\mathcal{I}$  & \textbf{61.57}$_{\color{red}{+1.87}}$   & \textbf{62.30}$_{\color{red}{+1.80}}$  \\  
\textbf{WS-FCN} (ours) + CRF & WideResNet-38 & $\mathcal{I}$  & \textbf{63.23}$_{\color{red}{+0.53}}$   & \textbf{64.22}$_{\color{red}{-0.08}}$ \\ 
\cdashline{1-5}[0.8pt/2pt]
\textbf{WS-FCN}$^\S$ (ours)  & WideResNet-38 & $\mathcal{I}$  & \textbf{64.02}$_{\color{red}{+4.32}}$  & * \\ 
\textbf{WS-FCN}$^\S$ (ours) + CRF & WideResNet-38 & $\mathcal{I}$  & \textbf{65.02}$_{\color{red}{+2.32}}$ & * \\ 
\end{tabular}
\label{tab4}}
\end{center}
\end{table}
\begin{table}[t]
\begin{center}
\renewcommand\arraystretch{1.6}
\setlength{\tabcolsep}{5pt}{
\caption{Comparisons to the state-of-the-art methods on the \emph{val} sets of MS COCO 2014~\cite{lin2014microsoft} in terms of mIoU ($\%$). ``Superv.'' is the corresponding training supervision ($\mathcal{I}$: image-level class label, $\mathcal{S}$: saliency masks).
``\S'' represents the results that we use ground-truth image-level class labels to remove masks of false positive classes predicted by our model. }
\begin{tabular}{ l c c | c } 
Methods & Backbone & Superv.  & \emph{val} (\%)\\ 
\hline \hline
\multicolumn{4}{l}{(\textbf{a}) \textbf{\emph{Multi-stage weakly-supervised}}} \\ 
\cdashline{1-4}[0.8pt/2pt]
DSRG~\cite{huang2018weakly} & VGG-16 & $\mathcal{I,S}$ & 26.00 \\
IAL~\cite{wang2020weakly} & VGG-16 & $\mathcal{I}$ & 27.70 \\
ADL~\cite{choe2020attention} & VGG-16 & $\mathcal{I}$  & 30.75 \\
SGAN~\cite{yao2020saliency} & VGG-16 & $\mathcal{I,S}$ & 33.60 \\ 
IRN~\cite{ahn2019weakly} & ResNet-50 & $\mathcal{I}$ & 32.60 \\
SEAM~\cite{wang2020self} & WideResNet-38 & $\mathcal{I}$  &  31.90 \\
CONTA~\cite{zhang2020causal} & WideResNet-38 & $\mathcal{I}$  &  32.80 \\
RIB~\cite{lee2021reducing} & ResNet-101 & $\mathcal{I}$  &  43.80 \\
\hline \hline
\multicolumn{4}{l}{(\textbf{b}) \textbf{\emph{Single-stage}}} \\ 
\cdashline{1-4}[0.8pt/2pt]
SSSS~\cite{araslanov2020single} & WideResNet-38 &  $\mathcal{I}$  & 26.68\\
SSSS~\cite{araslanov2020single} + CRF & WideResNet-38 &  $\mathcal{I}$  & 27.21\\
\cdashline{1-4}[0.8pt/2pt]
\textbf{WS-FCN} (ours) & WideResNet-38 & $\mathcal{I}$  & \textbf{27.34}$_{\color{red}{+0.66}}$ \\
\textbf{WS-FCN} (ours) + CRF & WideResNet-38 & $\mathcal{I}$  & \textbf{27.53}$_{\color{red}{+0.32}}$ \\
\cdashline{1-4}[0.8pt/2pt]
\textbf{WS-FCN}$^\S$ (ours) & WideResNet-38 & $\mathcal{I}$  & \textbf{32.65}$_{\color{red}{+5.97}}$ \\
\textbf{WS-FCN}$^\S$ (ours) + CRF & WideResNet-38 & $\mathcal{I}$  & \textbf{34.12}$_{\color{red}{+6.91}}$ 
\end{tabular}
\label{tab5}}
\end{center}
\vspace{-2mm}
\end{table}

\myparagraph{Quantitative results on MS COCO 2014~\cite{lin2014microsoft}.}
In this section, we conduct experiments on the much challenging MS COCO 2014~\cite{lin2014microsoft}. Table~\ref{tab5} shows results of our approach and other weakly-supervised methods~\cite{araslanov2020single,yao2020saliency,lee2021reducing,huang2018weakly,wang2020weakly} on the \emph{val} set. Compared to the existing SSSS model~\cite{araslanov2020single}, our \textbf{WS-FCN} outperforms it and achieves $27.53\%$ mIoU with the help of CRF. Undeniably, compared to some multi-stage methods, \eg, SGAN~\cite{yao2020saliency} and RIB~\cite{lee2021reducing}, our \textbf{WS-FCN} is not as good as them. Although SGAN~\cite{yao2020saliency} is $6.07\%$ mIoU higher than our proposed method, it is noting that it adopts saliency as additional supervision. Besides, \textbf{WS-FCN} is superior to DSRG~\cite{huang2018weakly}, which even adopts image-level class labels and saliency as supervisions. Meanwhile, the performance of our proposed method is close to IAL~\cite{wang2020weakly}, which has only a $0.17\%$ mIoU gap between them. It means single-stage methods can indeed simplify complicated procedures and outperform some multi-stage methods. 
We also provide better results on the condition that we remove the masks of false positive classes with ground-truth image-level class labels, which can achieve $34.12\%$ mIoU with the CRF operation. Overall, the results of COCO 2014 further indicate our proposed method can improve segmentation quality by alleviating the \textbf{\emph{background incompleteness}} and filling \textbf{\emph{object incompleteness}}.
\begin{figure*}[t]
\centering
\includegraphics[width=.99\textwidth]{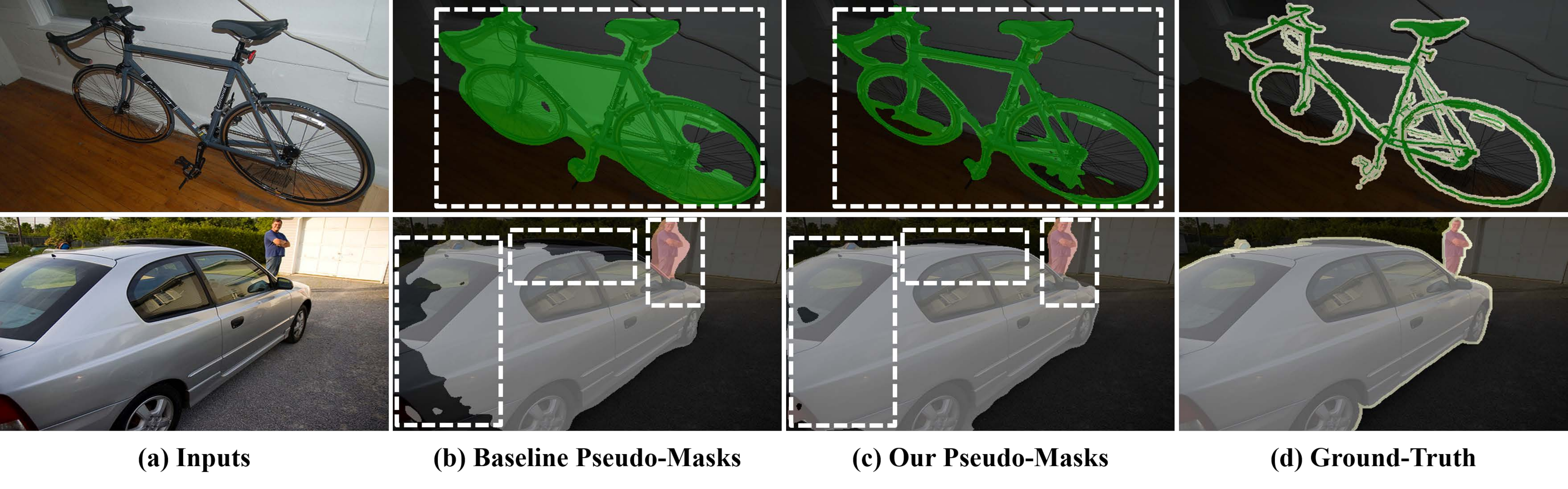}
\caption{Qualitative result comparisons on the \emph{val} set of PASCAL VOC 2012~\cite{everingham2010pascal}. The white dashed frames highlight the improved areas predicted by WS-FCN. It is worth noting that pixel-level ground-truth is only used for result evaluation, not in the model training process.}
\label{fig6}
\end{figure*}

\begin{figure*}[t]
\centering
\includegraphics[width=.99\textwidth]{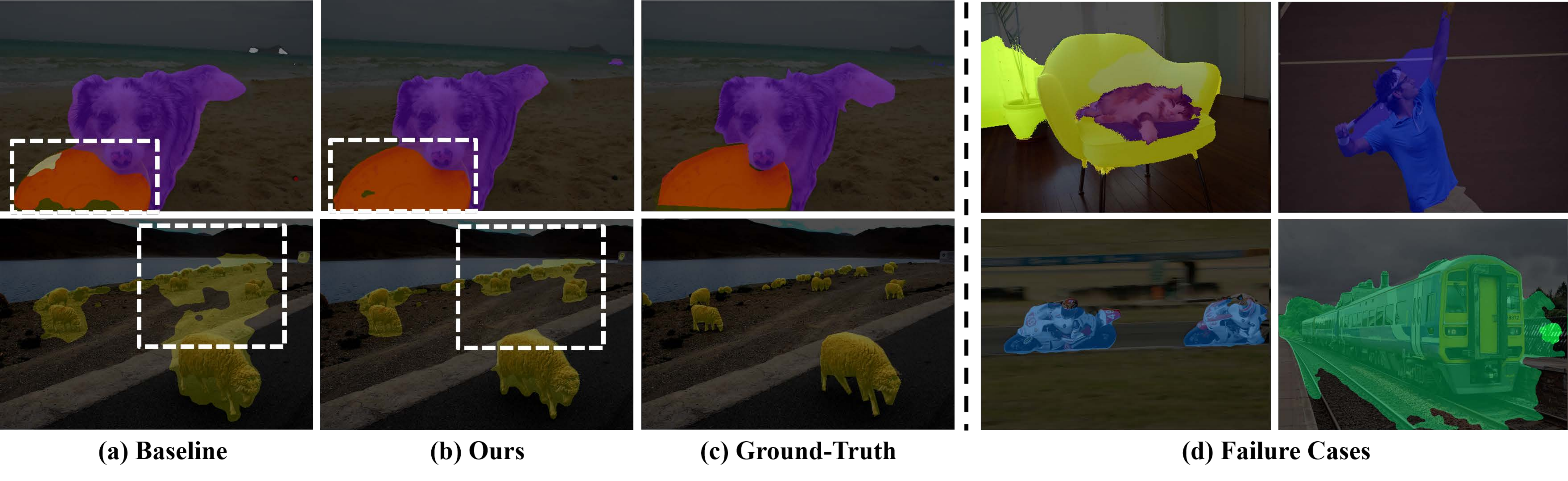}
\caption{Qualitative results on the \emph{val} set of MS COCO 2014~\cite{lin2014microsoft}.  We show the results of segmentation masks from baseline and our approach for comparison (left). At the same time, we exhibit the failure cases, CRF post-processing and ground-truth for visualization (right).}
\label{fig7}
\end{figure*}

\myparagraph{Qualitative results.} Figure~\ref{fig6} shows some visualization results of the baseline and our method on the \emph{val} set of PASCAL VOC 2012~\cite{everingham2010pascal}. Compared to the baseline~\cite{araslanov2020single} results, we can observe that our method can segment the objects more accurately. Concretely, \textbf{WS-FCN} can well fill the incomplete regions in the object, \eg, the ``\emph{car}". This is because our method brings sufficient spatial details into the high-level feature maps. Besides, we can also observe that \textbf{WS-FCN} can alleviate the problem of taking the background pixels as a part of the objects, \eg, the ``\emph{bicycle}". 
These results demonstrate that our method can capture the global context information.
Overall, the above visualization results can intuitively verify the effectiveness of our method. We also show some visualization results on the \emph{val} set of MS COCO 2014~\cite{lin2014microsoft} in Figure~\ref{fig7}.  Compared to baseline, we can see that our method can well deal with the mistakenly classified background pixels in the left side of Figure~\ref{fig7}, \eg, the ``\emph{sheep}". It not only validates our \textbf{WS-FCN} can capture abundant context information, but also demonstrates sufficient object context information is important to locate the boundaries of objects. In addition, our \textbf{WS-FCN} can complete the lack of object region, especially on the details of objects, \eg, the ``\emph{plate}". It can confirm the effectiveness of fusing the spatial details into deep feature maps and our method can integrate them well.

Besides, failure cases are shown in the right side of Figure~\ref{fig7}. We can observe that there are still imperfections in some special kind of object segmentation masks. For example, the cases of the first row in Figure~\ref{fig7} (d), {the ``\emph{legs}" of a chair and the ``\emph{tennis racket}" held up by the ``\emph{person}'' are hard to be accurately segmented due to the insufficient resolution representation from the backbone network, \ie, the output feature maps are $1/4$ of the input. Besides, some background stuff usually appears with the foreground objects together like the ``train" and the ``\emph{tracks}", the ``\emph{motorcyclist}" and the ``\emph{motorcycle}".} They are essentially a kind of spurious associations that background pixels can be easily taken as a part of the objects mistakenly, since these kinds of objects often appear together in the used dataset. This problem can be solved by removing spurious associations between different categories of objects through the causal intervention strategy.
\section{Conclusion}
In this work, we proposed a \textbf{WS-FCN} for single-stage weakly-supervised semantic segmentation, where two schemes termed as FCA and SF2 are used to address the problems of \emph{background incompleteness} and \emph{object incompleteness}, respectively. Through multi-scale feature aggregation and cross-layer feature fusion operations, the proposed WS-FCN can learn rich context information and make deep features have richer detailed information. Extensive experimental results on the challenging PASCAL VOC 2012 and MS COCO 2014 datasets demonstrated their effectiveness and efficiency. However, our method is still not able to deal well with too small objects or the co-occurrent problem between the background and the foreground. In the future, we will continue to work on more efficient solutions to these two problems. Besides, we will continue to introduce some extra-data multi-modal (\eg, sentences, and extra-images) into weakly supervised semantic segmentation to represent and distinguish them.
\bibliographystyle{IEEEtran}
\bibliography{IEEEtran}
\end{document}